\documentclass[sigconf]{acmart}
\pdfoutput=1

\usepackage{amsmath,amssymb, graphicx}
\usepackage[ruled,linesnumbered]{algorithm2e}
\usepackage{algorithmic}
\usepackage{paralist} 
\usepackage{color}
\usepackage{multirow}
\usepackage{rotating}
\usepackage[mathscr]{eucal} 
\usepackage{amsbsy} 
\usepackage{subfigure}
\usepackage{booktabs}
\usepackage{xcolor}
\usepackage{tikz}
\usetikzlibrary{snakes}
\usepackage{multirow}
\usepackage{balance}
\usepackage{tablefootnote}
\usepackage{empheq} 
\usepackage{moresize}
\usepackage{enumitem}
\setlist{nolistsep}

\setcopyright{rightsretained}
\acmDOI{10.475/123_4}
\acmISBN{123-4567-24-567/08/06}
\acmConference[KDD'18 Workshop]{ACM SIGKDD 2018}{}{London, UK}
\acmYear{2018}
\copyrightyear{2018}
\acmArticle{4}
\acmPrice{15.00}

\hypersetup{colorlinks,
    linkcolor=ACMRed,
    citecolor=ACMPurple,
    urlcolor=ACMDarkBlue,
    filecolor=ACMDarkBlue}

\newcommand{\hide}[1]{}

\newcommand{\ben}{\begin{enumerate*}}
\newcommand{\een}{\end{enumerate*}}
\newcommand{\bit}{\begin{itemize*}}
\newcommand{\eit}{\end{itemize*}}

\begin{document}

\author{Uday Singh Saini}
\affiliation{{University of California Riverside}
}
\email{usain001@ucr.edu}

\author{Evangelos E. Papalexakis}
\affiliation{{University of California Riverside}
}
\email{epapalex@cs.ucr.edu}

\title{A Peek Into the Hidden Layers of a Convolutional Neural Network Through a Factorization Lens}

\begin{abstract}
Despite their increasing popularity and success in a variety of supervised learning problems, deep neural networks are extremely hard to interpret and debug: Given an already trained deep neural network, and a set of test inputs, how can we gain insight into how those inputs interact with different layers of the neural network? Furthermore, can we characterize a given deep neural network based on its observed behavior on different inputs? In this paper, we propose a novel factorization-based approach on understanding how different deep neural networks operate. In our preliminary results, we identify fascinating patterns that link the factorization rank (typically used as a measure of interestingness in unsupervised data analysis) with how well or poorly the deep network has been trained. Finally, our proposed approach can help provide visual insights on how high-level, interpretable patterns of the network's input behave inside the hidden layers of the deep network. 

\end{abstract}

\maketitle
\keywords{}

\section{Introduction}
\label{sec:intro}

Deep neural networks have gained enormous popularity in machine learning and data science alike, and rightfully so, since they have demonstrated impeccable performance in a variety of supervised learning tasks , especially a number  of computer vision problems, most prominent examples being \cite{he2017maskrcnn},\cite{huang2017densely}. Albeit very successful in providing accurate classifications, deep neural networks are notorious for being hard to interpret, explain, and debug, a problem amplified by their increasing complexity. This is an extremely challenging problem and the jury is still out on whether it can be solved in its entirety.

Within the confines of interpreting and debugging deep neural networks, we are interested in answering the following questions: Given an already trained deep neural network, and a set of test inputs, how can we gain insight into how those inputs interact with different layers of the neural network? Furthermore, can we characterize a given deep neural network based on its observed behavior on different inputs? 

To the best of our knowledge, there is very limited prior work on the topic. One line of work that is attempting to answer such questions is the work by Bau et al. referred to as ``Network Dissection'' \cite{netdissect2017} and the work done by Olah, et al., "The Building Blocks of Interpretability", Distill, 2018, \cite{olah2018the}. Network Dissection is a framework which quantifies the interpretability of activations of hidden layers of CNNs. It does so by evaluating the alignment between neural activations  in the hidden units and a set of semantic concepts. In the work done on interpretability by Olah, et al., in "The Building Blocks of Interpretability", their focus is to learn what each neuron or a group of neurons detect based on feature visualization, and then attempts Spatial Attribution and Channel Attribution in order to explain how the network assembles these pieces to come at a decision. More recently, Raghu et al. \cite{raghu2017svcca} introduced a Canonical Correlation Analysis based study that jointly analyzes the hidden layers of a CNN, however, this analysis is not relating the derived representations of CCA to the input data, thus may not be able to provide an end-to-end characterization and visualization. 
Finally, a concurrent study to ours by Sedhi et al. \cite{sedghi2018singular} is focusing on analyzing the singular values of the convolutional layers of a CNN towards better regularization and quality improvement. To the best of our knowledge, our work is the first approach towards characterizing the quality of a CNN through a joint factorization lens.

In this work, we propose an  novel research direction that leverages factorization towards answering the above questions. The key idea behind our work is the following: we jointly factorize the raw inputs to the deep neural network and the outputs of each layer, to the same low-dimensional space. Intuitively, such a factorization will seek to identify commonalities in different parts of the raw input and how those are reflected and processed within the network. For instance, if we are dealing with a Deep Convolutional Neural Network that is classifying handwritten digits, such a joint latent factor will seek to identify different shapes that are common in a variety of input classes (e.g., round shapes for ``0'', ``6'', and ``9'') and identify potential correlation on how different layers behave collectively for such high-level latent shapes. 

This paper reports very preliminary work in that direction. The main contributions of this paper are:
\begin{itemize}[noitemsep]
	\item {\bf Novel problem formulation \& modeling}: We propose a novel problem formulation on providing insights into a deep neural network via a joint factorization model. 
	\item {\bf Experimental observations}: In three experimental case studies,  we train a Convolutional Neural Network in various problematic ways, and our proposed formulation reveals a persistent pattern that indicates a relation between the rank of the joint factorization and the quality training. {\em It is very important to note that those patterns are revealed without using labels for the test data.}
	\item {\bf Visualization Tool}: In addition to the link between the factorization rank and the training quality, our proposed method is able to provide visualizations that provide insights on how different high-level shapes/parts in the input data traverse the network.
\end{itemize}
\section{Proposed Method}
\label{sec:method}

\begin{figure}[!ht]
	\includegraphics[width=0.45\textwidth]{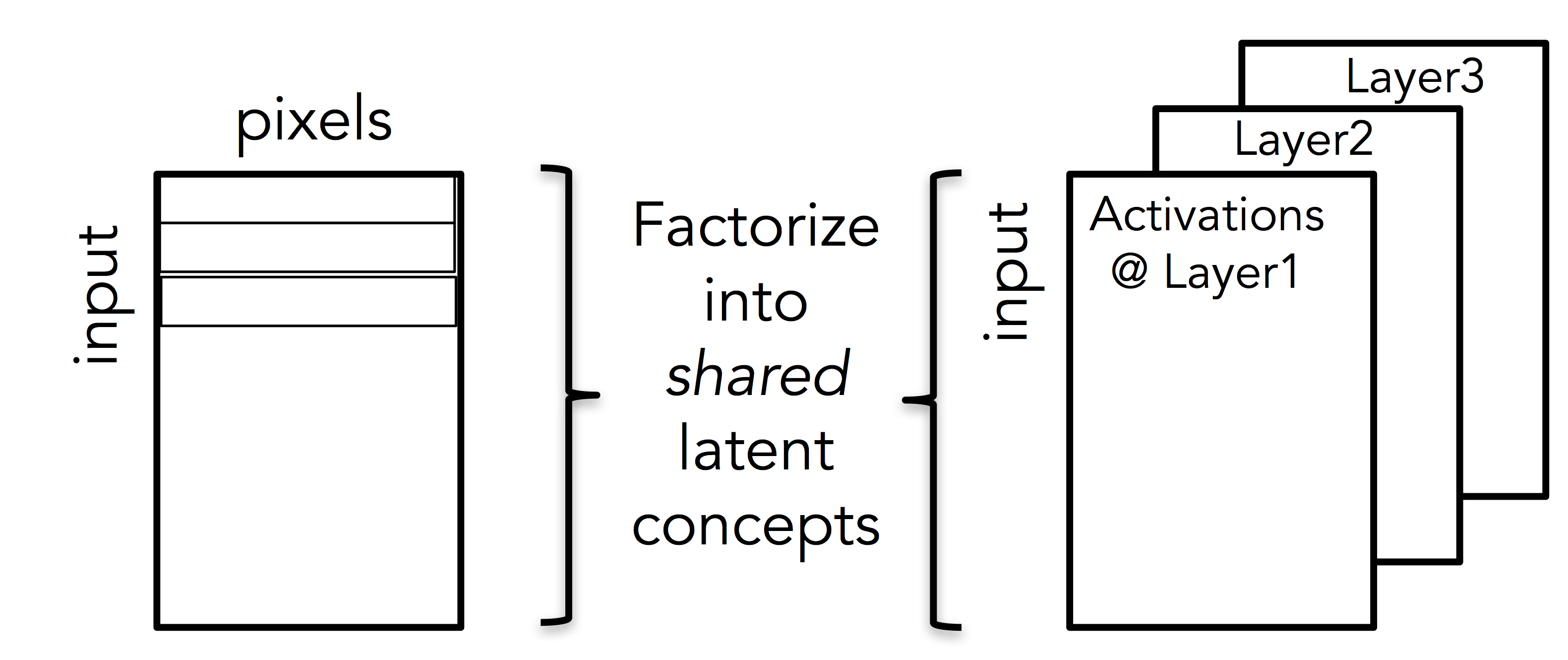}
	\caption{A high-level overview of our proposed modeling.}
	\label{fig:model}
\end{figure}

As mentioned in the introduction, given an already trained neural network and a set of test data (without their labels), we seek to factorize the input data and the output of each hidden layer for the same data, into a joint low-dimensional space. A high-level overview of our proposed modeling is shown in Figure \ref{fig:model}.

The above formulation can be seen as capturing the joint latent factors that characterize the input data and the non-linear transformations thereof through the deep neural network.

In the following lines we provide details of our model and the fitting algorithm.

\subsection{Model Details}
Objective Function for Coupled Non-negative Factorization is as follows:\\
 \begin{equation}\label{eq:1}\begin{split}J(\textbf{P},F,\textbf{O}) = \sum_{i = 0}^{C-1} \lVert D_i - P_iF^{T}\rVert_{F}^{2}   +  \sum_{j = 0}^{L-1} \lVert A_j - O_jF^{T}\rVert_{F}^{2}  \\ \ni P_{i}, O_{j}, F, \ are \ element \ wise \geq 0 \forall \ i,j \end{split}\end{equation},
where C is the number of channels in an input image and L is the number of layers of the neural network being analysed, \textbf{P} and \textbf{O} are sets of matrices $ \{P_i : \forall 0 \leq i \leq C-1 \in \mathbb{Z}\}$ and $ \{O_j : \forall 0 \leq j \leq L-1 \in \mathbb{Z}\}$ respectively. Each $D_i$ is the set of $i^{th}$ channel of the input images to the neural network, where each column of $D_i$ is a channel of the image in vectorized form, thus each row of $D_i$  is a pixel or location in the original image. For Grayscale images, the number of channels is 1, hence in such a scenario each column of $D_0$ represents an input image fed to the neural network. Similarly, each $A_j$ is the matrix of activations of the $j^{th}$ layer of the neural network, where each column of $A_j$, for instance, the $k^{th}$ column $A_j[:,k]$, is the activation of layer $j$ of the network for $k^{th}$ input image, $i^{th}$ channel of which is represented by $D_i[:,k]$.\\
Each $P_i$ is a matrix that stores the latent representation of each pixel (for the $i^{th}$ channel)  in it's rows, each $O_{j}$ is a matrix that stores the latent representation of each neuron (or activation) of layer $j$ in it's rows. Finally, $F$ is a matrix that stores the latent representation of each image fed to the neural network  in it's rows.\\
The first summation of the objective function is geared at finding structures at pixel-level in the input images (for all channels in the input image) to the network. The second summation term tries to find patterns between neural activations for various inputs. The coupling matrix $F$ propagates information between the 2 summations and our goal in doing so is to infer correlations or patterns between clustering of input images and clustering of neurons (across all layers),i.e., we aim to investigate whether the same cluster neurons fire for similar (yet not identical) inputs. This approach is inspired by broader goal of understanding the relation between the discriminative power of neural networks vs their interpret-ability. \\
We solve equation \eqref{eq:1} using the algorithm provided in \cite{NIPS2000_1861} and the update steps are as follows:-\\

\begin{eqnarray*}
  F &\leftarrow& F * \frac{\sum\limits_i D_{i}^TP_{i} +\sum\limits_j A_{j}^TO_{j}}{\sum\limits_i FP_{i}^TP_{i} +\sum\limits_j FO_{j}^TO_{j}} \\
  P_i &\leftarrow& P_i * \frac{D_{i}F}{P_iF^TF} \ \  \forall i\\
 O_j &\leftarrow& O_j * \frac{A_jF}{O_jF^TF} \ \ \forall j\\
\end{eqnarray*}
where  $*$ stands for Hadamard (element  wise) product\footnote{$(A * B)_{i,j} = A_{i,j} \times B_{i,j}$} and $/$ stands for element wise division \footnote{$(A / B)_{i,j} = A_{i,j} \div B_{i,j}$}. For numerical stability, a small constant $\epsilon$ is added element-wise to the resultant matrix in the denominator. We initialize $F$, $P_i$'s, $O_j$'s randomly with component values between 0 and 1.

The model above is a proof-of-concept and is not leveraging higher-order correlations in the input and the activations layers; such higher-order dynamics can be exploited via tensor modeling which is an active direction of extending this model.

\section{Experimental Analysis}
\label{sec:experiments}
In this section we present our analysis of the neural network via our coupled Non-negative Factorization framework. We proceed as follows:
\begin{itemize}
 \item We first provide details about the experimental setup: Dataset and the Neural Network.\item We describe how we setup our model for analysis of the network.
 \item Next we try to study the behavior of the network on a fixed test set with respect to variations in the amount of training data via our model \eqref{eq:1}.
 \item We study similar behaviour as above, though this time with we train the network only on a subset of categories.
\end{itemize}
\subsection{Dataset and The Network}
We used a raw MNIST Dataset\footnote{\url{https://github.com/myleott/mnist_png}} which about 60,000 training images and 6,000 test images. Each image is a single channel gray-scale with a resolution of 28 by 28 pixels.\\
The network we analyze consists of 2 consecutive Convolutional Layers with Maxpool and ReLU, followed by a fully connected layer which feeds to a softmax output. For our study we focus on analyzing the 2 convolutional layers. The first convolutional layer has 1 input channel and yields 10 output channels with a kernel size of $5 * 5$, which leads  to a maxpool and subsequently to a ReLU output. This output is fed to the second convolutional layer which takes 10 input channels and outputs 20 channels, again with a kernel size of $5 * 5$ with a subsequent maxpool and ReLU. We refer to the output of ReLU as the activations for that Layer, given the input.

\subsection{Setting up the model }
In this section we describe how we construct the matrices $D_i$'s and $A_j$'s. For simplicity we consider only grayscale or single-color channel input images. We take an input image  that is fed to the network, we vectorize it, and stack in a column of  $D_0$, Note that the suffix is 0 since the input is a single channel image. Thus, when we take the $k^{th}$ input image fed to the network, we vectorize it and set the  $k^{th}$ column of $D_0$, i.e. , $D_0[:,k]$ equal to the vectorized form of that image. For this image, we vectorize the activations of the $j^{th}$ layer, then we store the vectorized activations of the $j^{th}$ layer for the $k^{th}$ image in $A_j[:,k]$ $\forall \ j$. We repeat this process for all images in the test set.  If the input image is of size m x n x $C$ (single channel), and the number of images in the test set is $T$, then $D_i$ $\epsilon \mathbb{R}^{mn * T}$ $\forall \ 0 \leq i \leq C-1$. Let $N_j$ be the number of neurons in layer $j$ of the network. Then $A_j$ $\epsilon  \mathbb{R}^{N_j * T}$ $\forall \ j$. \\

\begin{figure*}[!ht]
	\subfigure[9\label{fig:9latent}]{\includegraphics[scale=0.2]{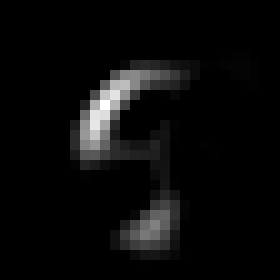}}
    \subfigure[1\label{fig:1latent}]{\includegraphics[scale=0.2]{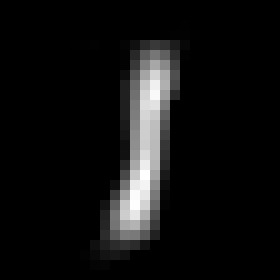}}
    \subfigure[6\label{fig:6latent}]{\includegraphics[scale=0.2]{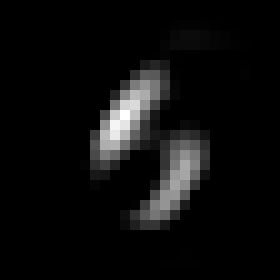}}
    \subfigure[0\label{fig:0latent}]{\includegraphics[scale=0.2]{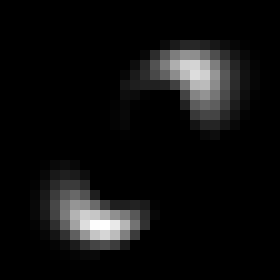}}
    \subfigure[3 and 8\label{fig:3latent}]{\includegraphics[scale=0.2]{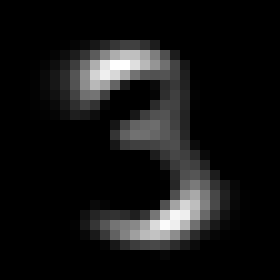}}
    \subfigure[7\label{fig:7latent}]{\includegraphics[scale=0.2]{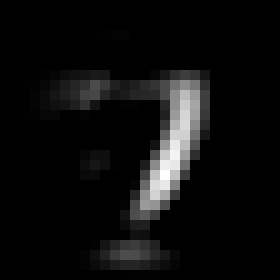}}
    \subfigure[1 and 2\label{fig:21latent}]{\includegraphics[scale=0.2]{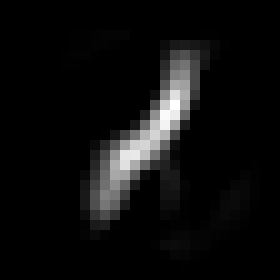}}
    \caption{Latent images and their corresponding class(es) on the MNIST dataset}
\end{figure*}

\subsection{Case Study I}\label{ssec:EE1}
In this section we describe the behavior of the network when we provide it with increasing amounts of training data, thereby improving it's performance on our test set. In this evaluation exercise, we train the network on dataset size varying from 25\% to 100\% in increments of 25\%. The accuracy on the test set for each sample is is as follows - for 25\% data: 83\%. for 50\% data: 89\%, for 75\% data: 93\%, for 100\% data: 95\%. For each sample size, we only train over 1 epoch of the data to maintain uniformity. We store the test set and the activations of the network over the test set in $D_i$'s and $A_j$'s, respectively, as explained earlier. We run the coupled non-negative factorization model once we obtain $D_i$'s and $A_j$'s for a particular instance of the experimental exercise. The number of latent factors in \eqref{eq:1} is varied from 10 to 50 with increments of 10.  

\begin{figure}[!htb]
\centering
\includegraphics[scale=0.5]{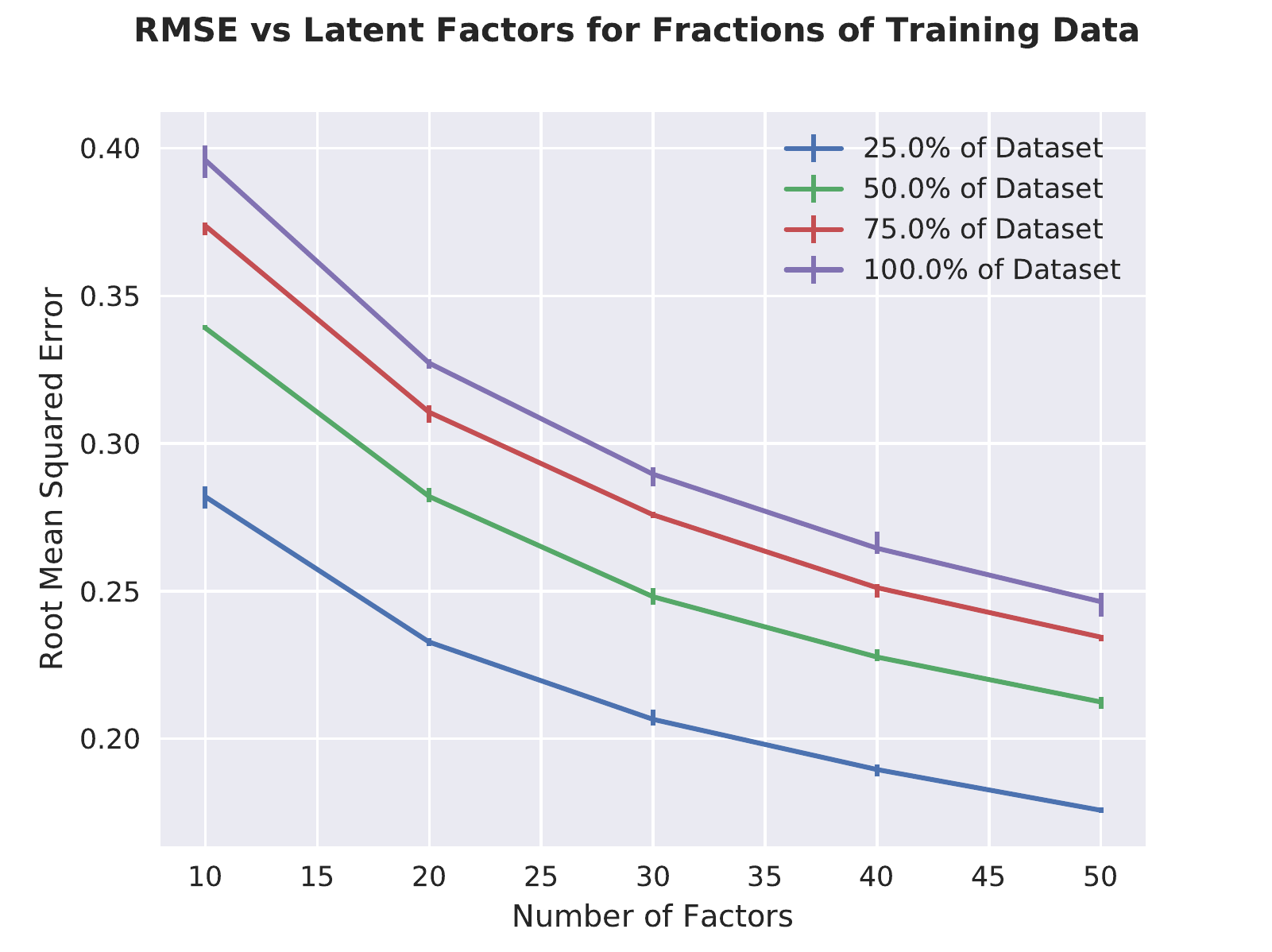}
\caption{RMSE of Objective function as the Training Data of the network is varied.}
\label{fig:RMSESample}
\end{figure}

\begin{figure}[!htb]
\centering
\includegraphics[scale=0.5]{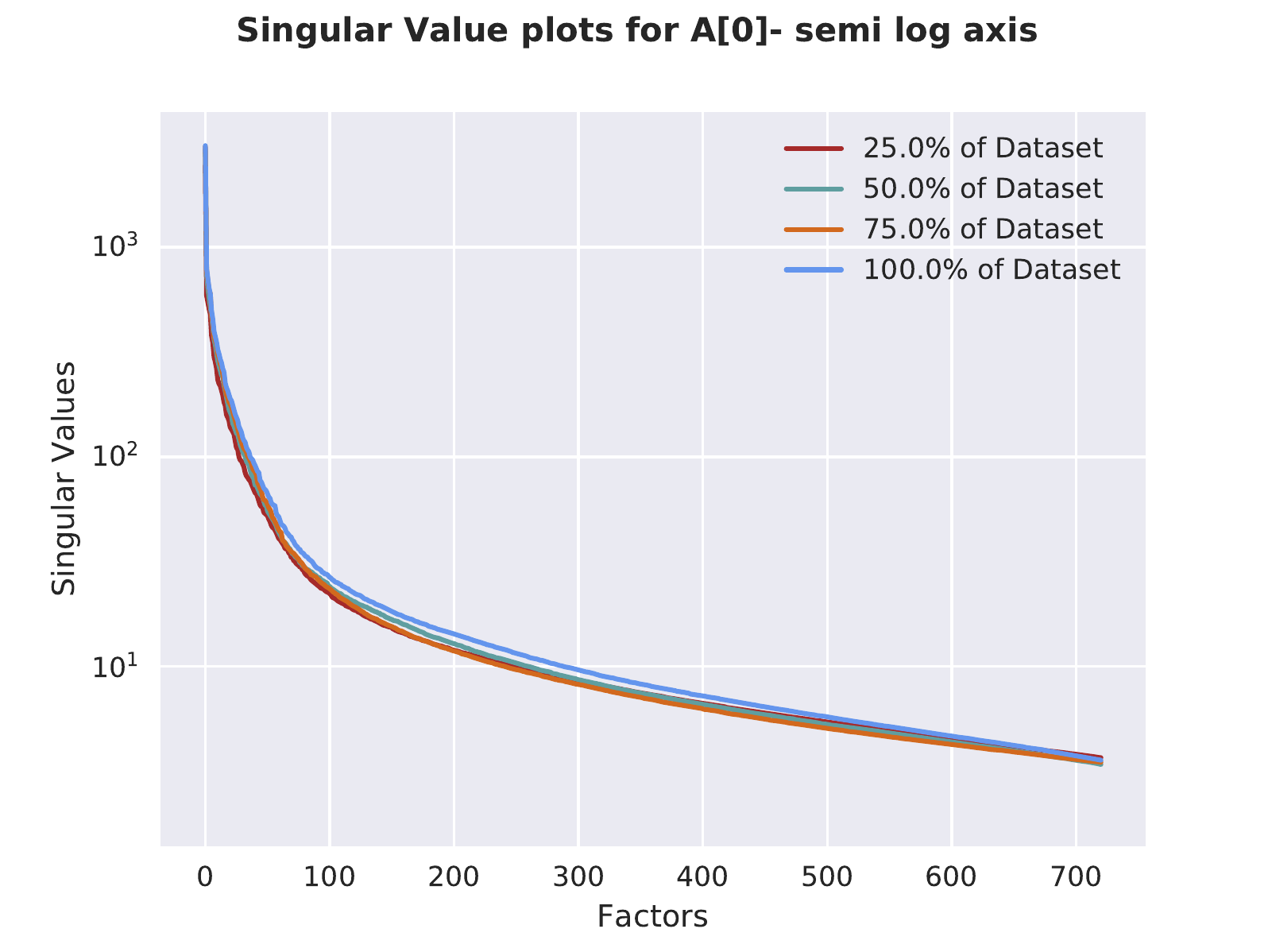}
\caption{Semi-log axis plot of top singular values of $A_0$.}
\label{fig:SVA0}
\end{figure}

\begin{figure}[!htb]
\centering
\includegraphics[scale=0.5]{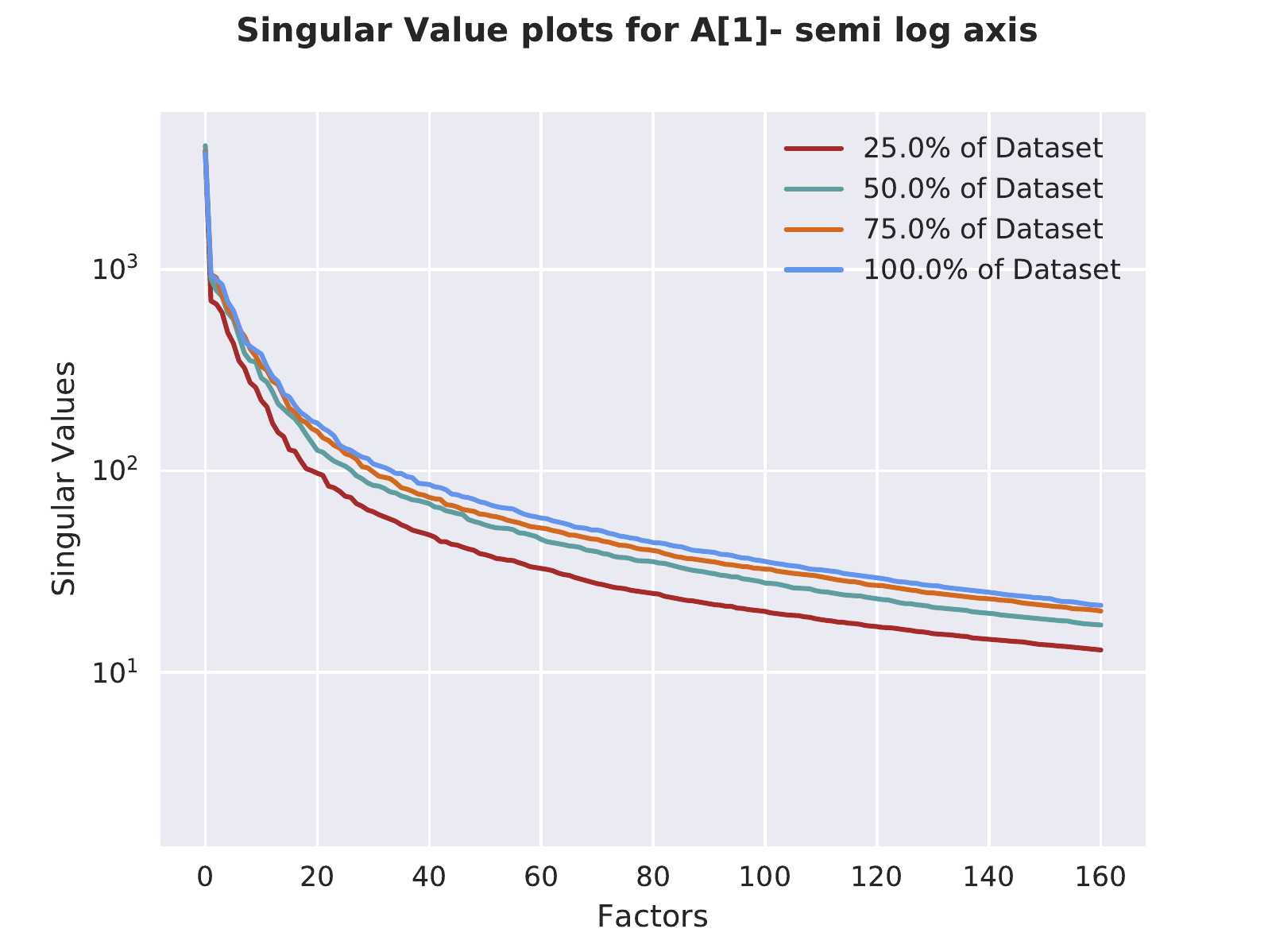}
\caption{Semi-log axis plot of top singular values of $A_1$.}
\label{fig:SVA1}
\end{figure}

The results are tabulated in \autoref{fig:RMSESample}. We observe that the outputs of the network when trained on smaller datasets tends to be more compressible, i.e., it requires a lower number of latent dimensions to explain itself, as evinced by the lower RMSE for smaller datasets over all latent dimension values. This understanding is further emboldened when we look at top singular values of all Activation matrices, $A_j$'s of the network under different training scenarios in \autoref{fig:SVA0} and \autoref{fig:SVA1}. Especially when we look at singular value plots given by \autoref{fig:SVA1} of the deepest layer of the network, we observe a clear difference between the singular value spectra in various training scenarios. Usually the final layers of a network are usually Fully connected layers followed by softmax or sigmoid non-linearities, and the goal of the previous layers is to non linearly project the input vector into a space where vectors belonging to various classes are easily separable by applying a fully connected layer with softmax. It becomes amply clear that a network with poorer performance transforms the dataset into a much lower dimensional subspace when compared with a well trained or a high-performance network. We would like to emphasize that if this assertion is accurate and omnipresent, our model doesn't need test data annotations to investigate relative performance of neural networks.

\begin{figure}[!htb]
\centering
\includegraphics[scale=0.5]{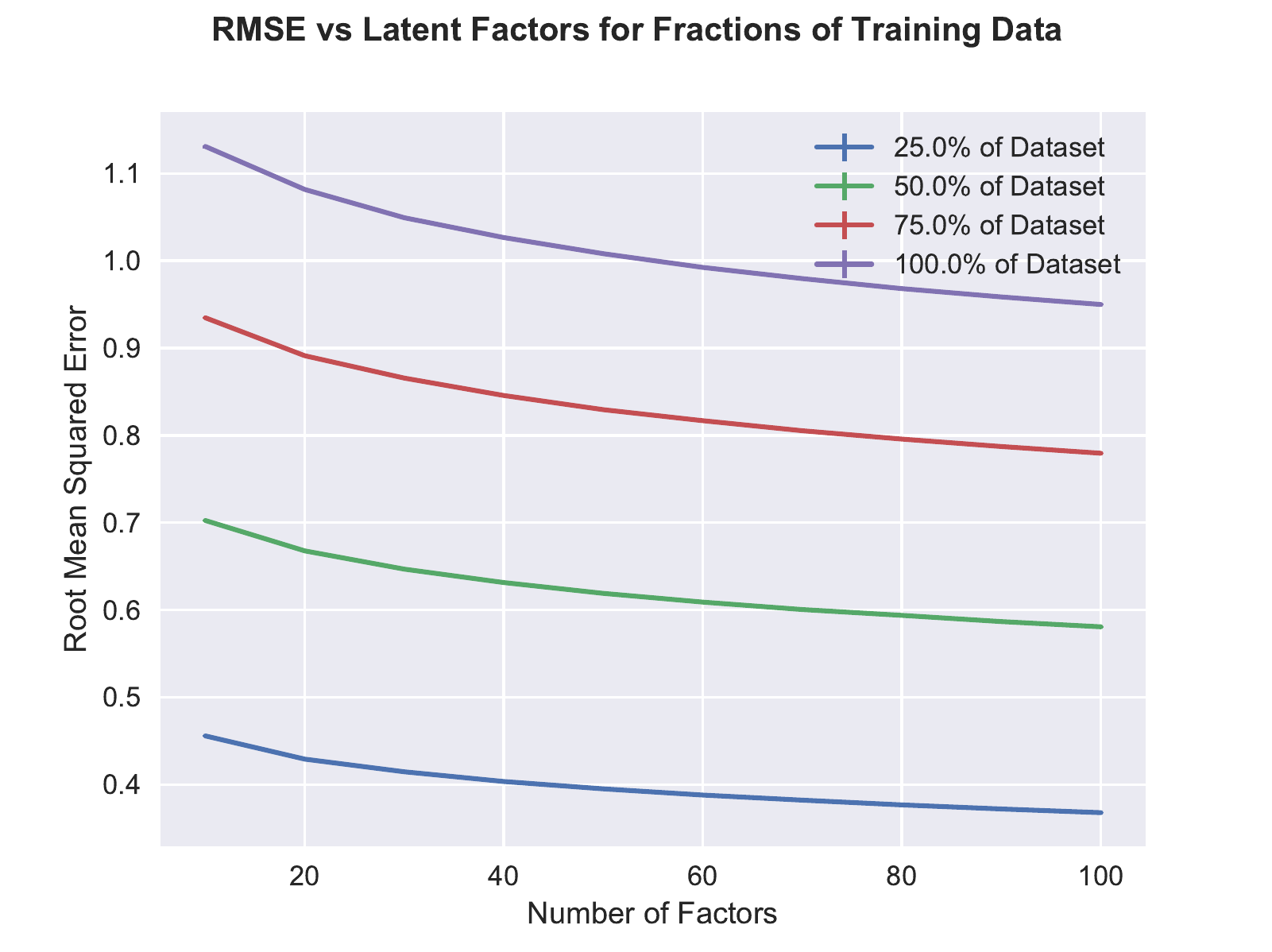}
\caption{RMSE of Objective function as the Training Data of the network is varied for CIFAR-10 using AlexNet}
\label{fig:RMSESampleAlexNet}
\end{figure}

In order to investigate the generality of the observed pattern, we also applied our factorization to a modified version of AlexNet \cite{krizhevsky2012imagenet} \footnote{We obtained the modified architecture definition from \url{https://github.com/bearpaw/pytorch-classification}} (where the modification was due to changing the size of the input) both on the MNIST and CIFAR-10 datasets, and the observations persisted. Figure \ref{fig:RMSESampleAlexNet} shows the RMSE for varying percentages of the training input for CIFAR-10 on AlexNet.

\begin{figure}[!htb]
\centering
\includegraphics[scale=0.5]{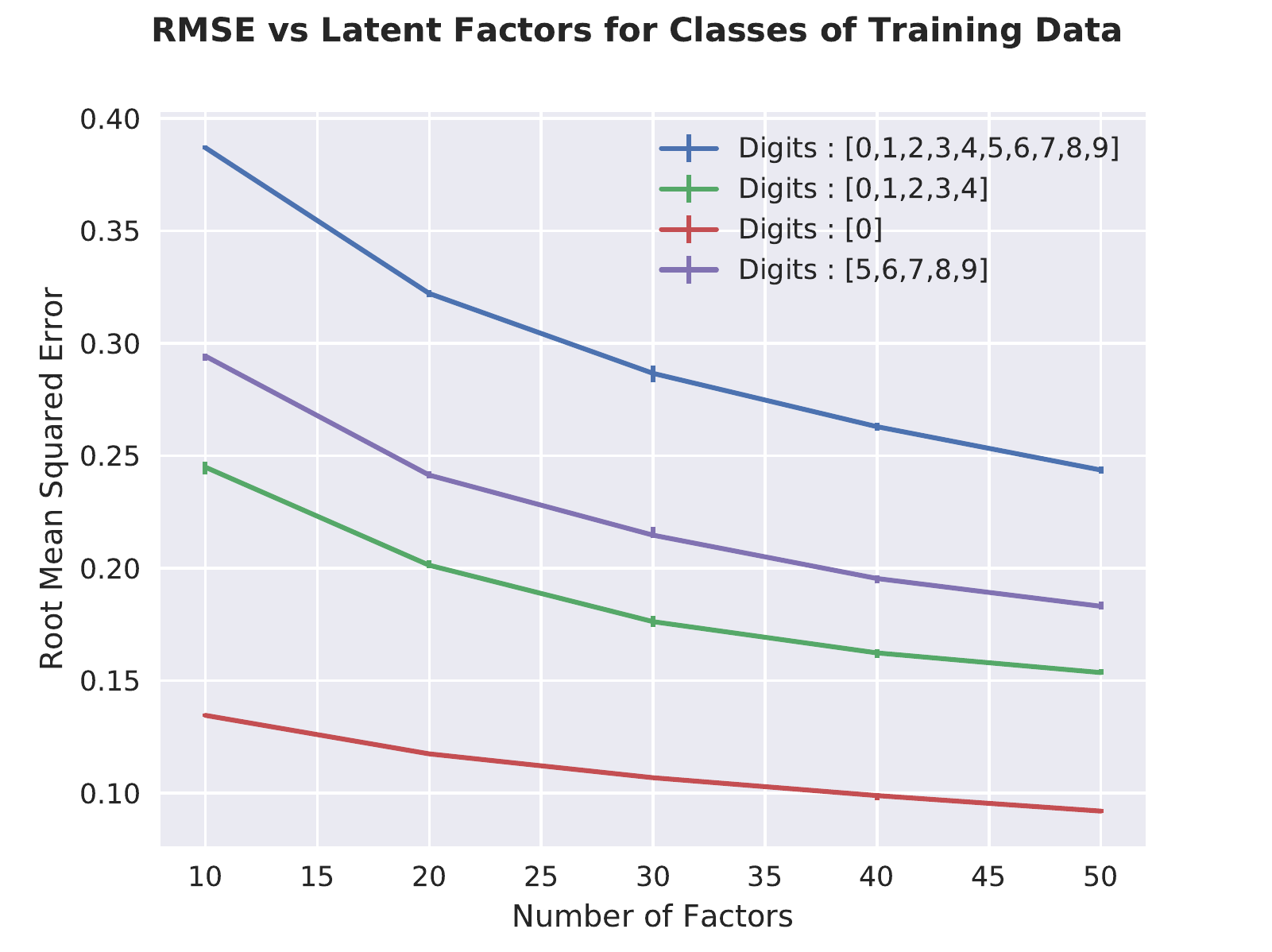}
\caption{RMSE of Objective function as the Subset of Training Classes is varied.}
\label{fig:RMSEClass}
\end{figure}

\subsection{Case Study II}
\begin{figure}[!htb]
\centering
\includegraphics[scale=0.5]{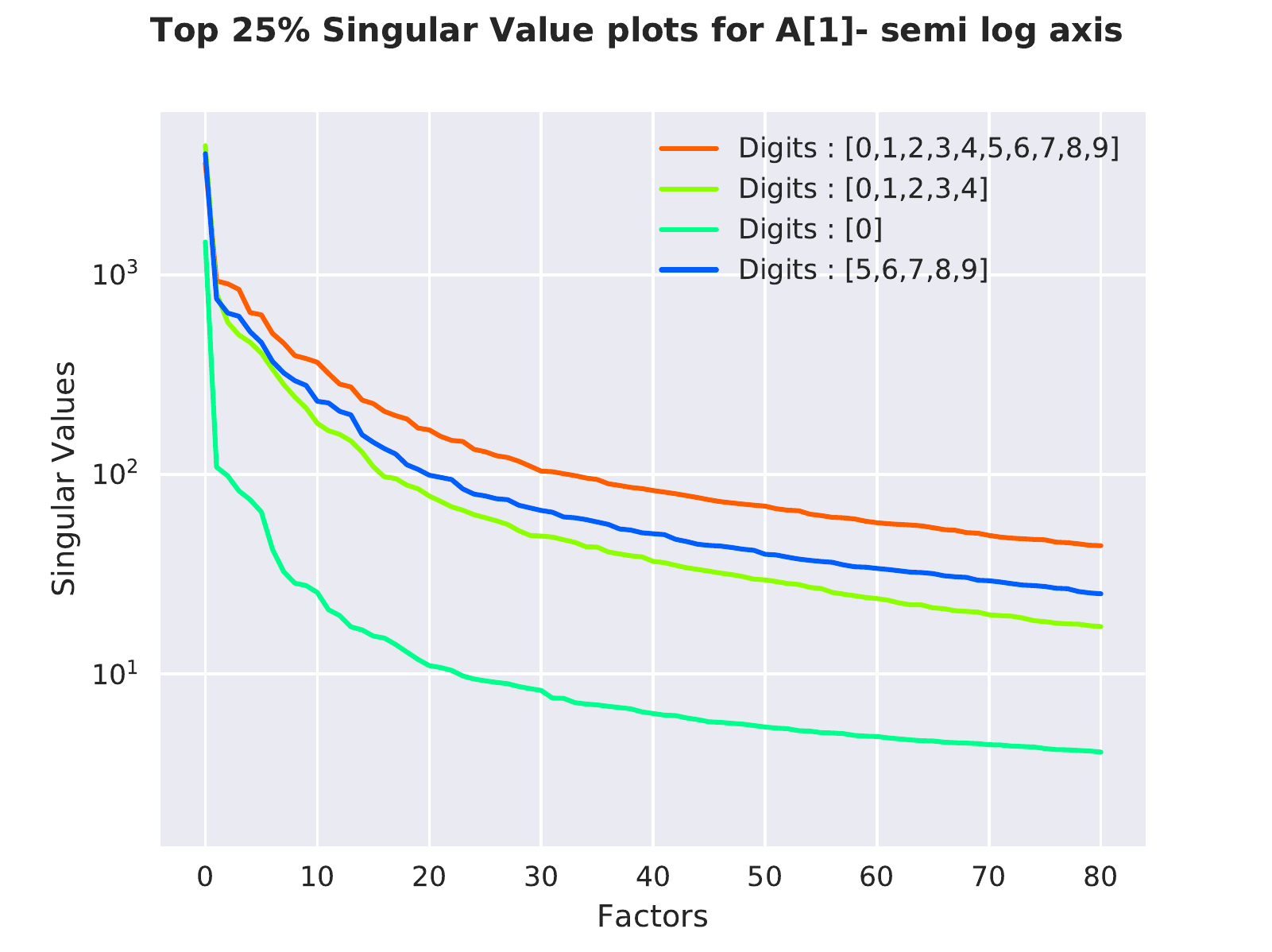}
\caption{Semi-log axis plot of top singular values of $A_1$ when the neural network is trained on various subsets of classes}
\label{fig:RMSEClassSVA1}
\end{figure}

In this section we describe the behavior of the network when we provide it with a subset of original classes of the training data. We train the network on the following subsets of digits \{0\},\{0,1,2,3,4\},\{5,6,7,8,9\} and \{0,1,2,3,4,5,6,7,8,9\}. The test set accuracies for the respective cases are 9\%,50\%,47\% and 93\%. The number of training examples for digits 0 through 4 were slightly more compared to the case of digits 5 through 9, hence the slight variation in accuracy.  For each subset, we only train over 1 epoch of the relevant data to maintain uniformity. As explained earlier, we store the test set and the activations of the network over the test set in $D_i$'s and $A_j$'s, respectively and rest of the setup is same as in \autoref{ssec:EE1}. The results of our model's analysis on this training setup are shown in \autoref{fig:RMSEClass}. We can clearly see that when the training data is small and/or only a class based subset of the original training data, the outputs of the network are much more compressible, as indicated by the lower RMSE for respective cases. This observation is even further consolidated by evidence from the singular value spectra (\autoref{fig:RMSEClassSVA1}) of the final activation layer of the network when trained with different subsets of classes.

\subsection{Case Study III}
In this study, we train the network in such a way that the input training examples are not sent in an arbitrary order, but instead, they are fed on a class by class basis. To give a hypothetical example, initially all the images for Digit 1 are given as input to the network for training, then all images for Digit 2, and so on. For the purpose of this study, the first training class was the Digit 9. The accuracy of the network on the test set was $9\%$, the network correctly recognizes most of the $9's$. As before, this is done only for 1 epoch over the dataset.\\
What makes this study interesting from our point of view is the fact that though the network has been trained on the entire dataset, but having been trained in such an orderly fashion, it is only good at recognizing the first class it was trained on, it would be interesting to see which neurons fire for other digits, and whether there are any pattern (similarities or contrasts) between the firing of neurons of a layer for various input digits. \\
During our experimentation with the MNIST dataset under this setting visualizations for the final convolutional layer in one of the test setup yielded the following results:-

\begin{figure}[!ht]
	\subfigure[Layer 1]{\includegraphics[width = 0.23\textwidth]{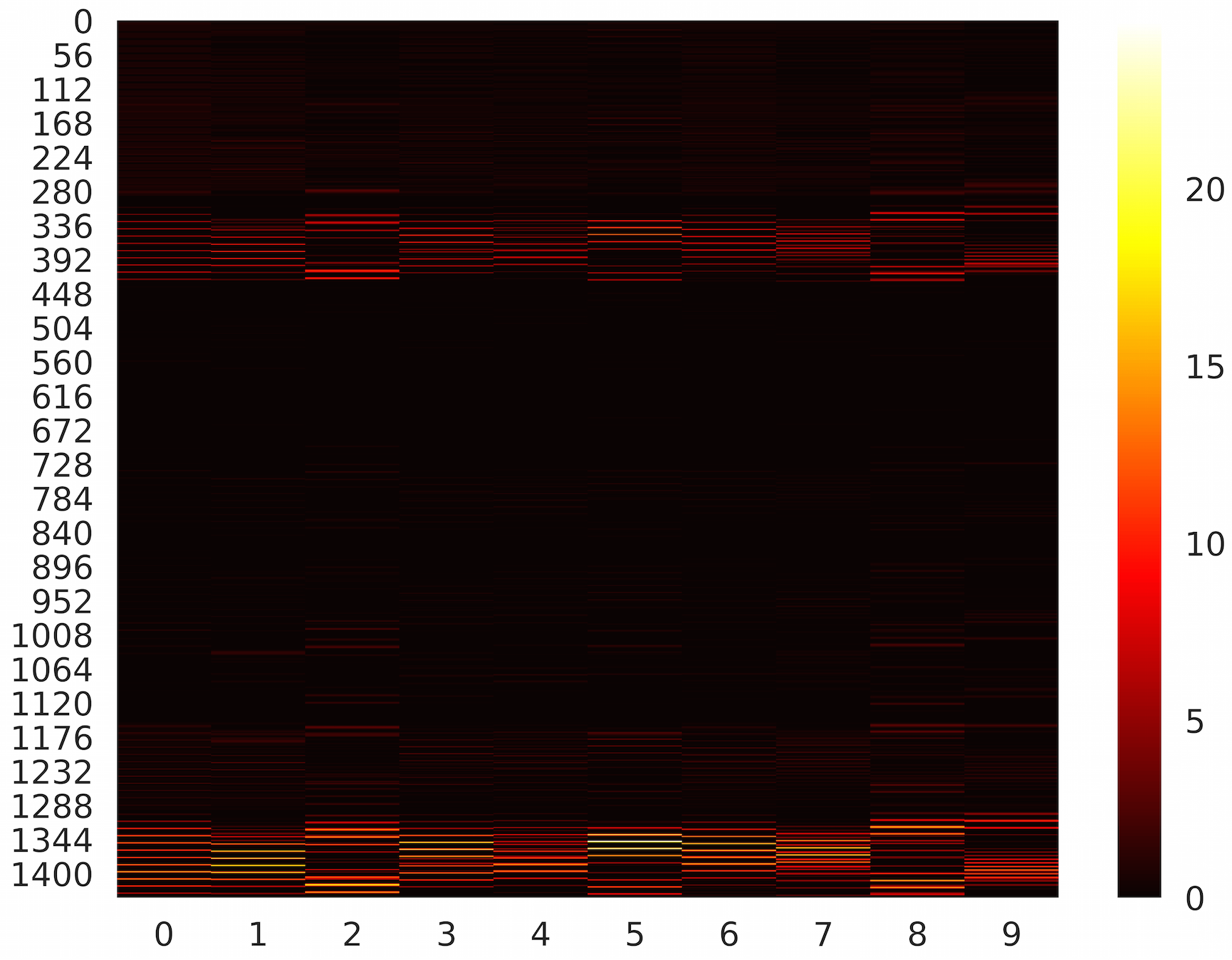}}
    \subfigure[Layer 2]{\includegraphics[width = 0.23\textwidth]{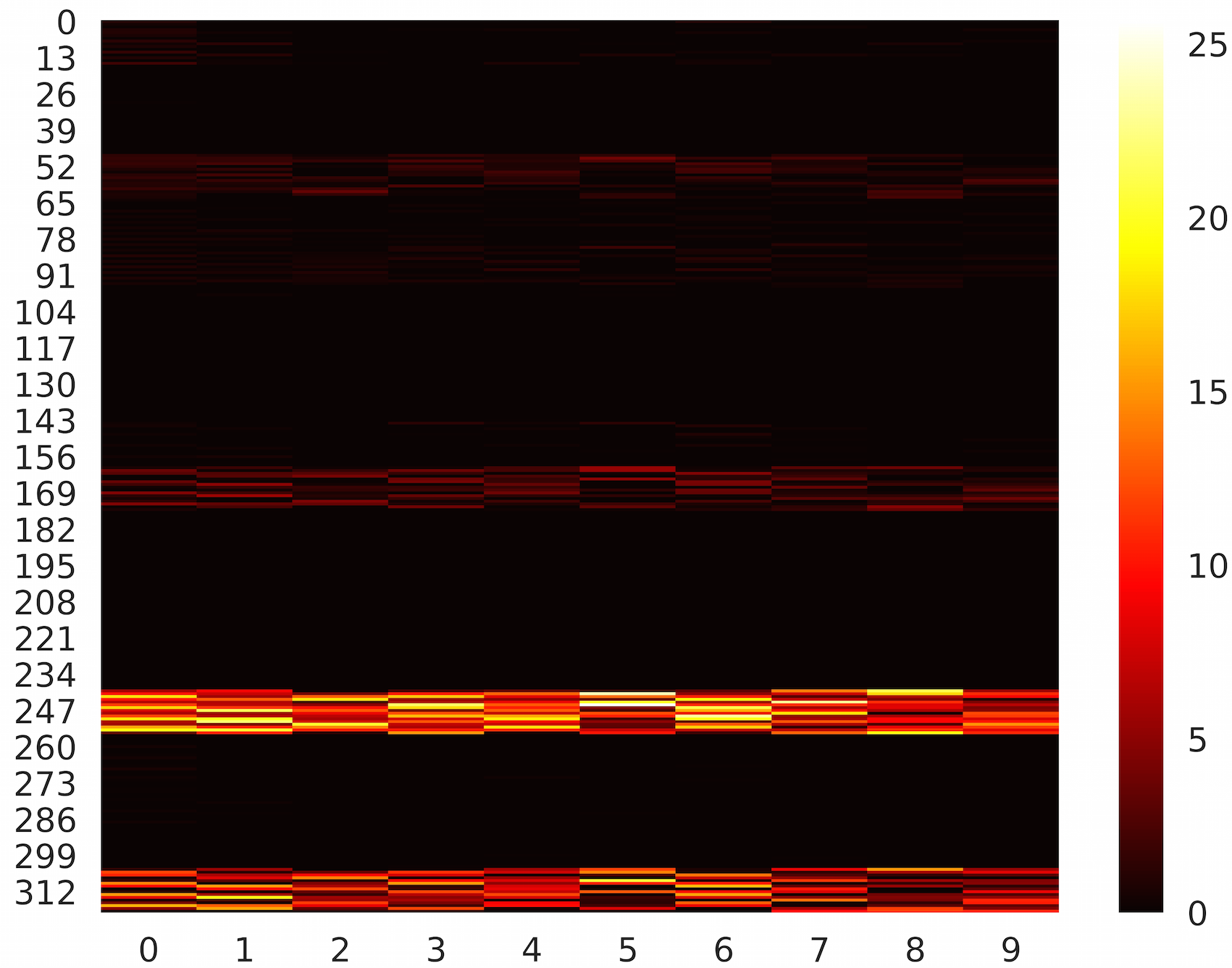}}
    \caption{Heatmap of the neurons of each layer with respect to their participation in each latent factor for a 10-component factorization, when the CNN was trained without shuffling. We observe that the network is heavily underutilized, since the latent patterns of neurons are limited, further corroborating our low-rank observation.}
    \label{fig:layer_viz}
\end{figure}

\begin{enumerate}
\item First, Considering the case on which the network performs well, i.e., when the input is digit $9$, the Neurons in the final layer which were active when the input was digit $9$ (See \autoref{fig:9latent}), were also active when the input images were digits $1$ and $6$ (See \autoref{fig:1latent} and \autoref{fig:6latent} respectively) . Indicating a commonality of structure among these digits. 

\item Among the examples on which the network performed poorly. We found that the neurons which fire for Digit $1$ (See \autoref{fig:1latent}), also fire for digits $7$ (See \autoref{fig:7latent}) and digits $9$ (See \autoref{fig:9latent}). Further, similar behavior was observed for  digits $7$ (See \autoref{fig:7latent}) and $2$ ( \autoref{fig:21latent}).

\item Turning our eye to the cases where the disjointness in the sets of firing of neurons for various groups of digits. We observe that the neurons which fire for  $0$ (See \autoref{fig:0latent}) are different from the neurons which fire for digits: $3$ and $8$ (See \autoref{fig:3latent}).
\end{enumerate}

Finally, in \autoref{fig:layer_viz} we show the latent factor heatmap for each neuron, for both layers of our CNN, for a rank 10 factorization. Even by a quick glance at the heatmaps, it becomes apparent that most of the network is not properly utilized in the case where we do not shuffle during training (which further corroborates our low-rank observation). 
We reserve further investigation of the visualization capabilities of our formulation for future work.

\section{Conclusions}
\label{sec:conclusions}
In this paper, we introduce a novel factorization-based method for providing insights into a Deep Convolutional Neural Network. In three experimental case studies, we identify a prominent pattern that links the rank of the factorization, roughly a measure of the degree of ``interestingness'' in a high-dimensional dataset, and the quality with which the network was trained: the poorer the training, the lower the rank. Importantly, this observation is derived in the absence of test labels. We intend to further investigate whether this observation holds in a wide variety of cases, and what other implications that would entail. Finally, we provide a visualization tool that helps shed light into how different cohesive high-level patterns in the input data traverse the hidden layers of the network.

\section{Acknowledgements}
{
The authors would like to thank NVIDIA for a GPU grant which facilitated computations in this work.
}

\balance
\bibliographystyle{ACM-Reference-Format}
\bibliography{vagelis_refs.bib}

\end{document}